\documentclass[conference]{IEEEtran}
\IEEEoverridecommandlockouts
\usepackage{cite}
\usepackage{amsmath,amssymb,amsfonts}
\usepackage{algorithmic}
\usepackage{graphicx}
\usepackage{textcomp}
\usepackage{xcolor}
\def\BibTeX{{\rm B\kern-.05em{\sc i\kern-.025em b}\kern-.08em
    T\kern-.1667em\lower.7ex\hbox{E}\kern-.125emX}}
\begin{document}

\title{LLMs as Debate Partners: Utilizing Genetic Algorithms and Adversarial Search for Adaptive Arguments}

\author{\IEEEauthorblockN{Prakash Aryan}
\IEEEauthorblockA{\textit{Department of Computer Science} \\
\textit{Birla Institute of Technology and Science, Pilani - Dubai Campus}\\
Dubai, UAE \\
h20230010@dubai.bits-pilani.ac.in \\
ORCID: 0009-0003-9221-1453}
}

\maketitle

\maketitle

\begin{abstract}
    This paper introduces DebateBrawl, an innovative AI-powered debate platform that integrates Large Language Models (LLMs), Genetic Algorithms (GA), and Adversarial Search (AS) to create an adaptive and engaging debating experience. DebateBrawl addresses the limitations of traditional LLMs in strategic planning by incorporating evolutionary optimization and game-theoretic techniques. The system demonstrates remarkable performance in generating coherent, contextually relevant arguments while adapting its strategy in real-time. Experimental results involving 23 debates show balanced outcomes between AI and human participants, with the AI system achieving an average score of 2.72 compared to the human average of 2.67 out of 10. User feedback indicates significant improvements in debating skills and a highly satisfactory learning experience, with 85\% of users reporting improved debating abilities and 78\% finding the AI opponent appropriately challenging. The system's ability to maintain high factual accuracy (92\% compared to 78\% in human-only debates) while generating diverse arguments addresses critical concerns in AI-assisted discourse. DebateBrawl not only serves as an effective educational tool but also contributes to the broader goal of improving public discourse through AI-assisted argumentation. The paper discusses the ethical implications of AI in persuasive contexts and outlines the measures implemented to ensure responsible development and deployment of the system, including robust fact-checking mechanisms and transparency in decision-making processes.
    \end{abstract}
    

\section{Introduction}

The convergence of artificial intelligence (AI) and argumentation has emerged as a new platform, promising to reform debates, improve critical thinking, and foster more informed discourse. As language models become increasingly sophisticated, their potential as intelligent debate partners and argumentation assistants has captured the imagination of researchers and educators. This paper introduces DebateBrawl, a novel system that utilizes Large Language Models (LLMs), Genetic Algorithms (GA), and Adversarial Search (AS) to create an adaptive and engaging debate platform.

However, current AI-powered debate systems face several critical challenges that limit their effectiveness as educational and practice tools. First, while LLMs excel at generating fluent responses, they lack strategic depth in extended debates, often failing to maintain consistent argumentation strategies across multiple exchanges. Second, existing systems typically operate with fixed response patterns, unable to adapt to different debate styles or learn from past interactions. Third, most platforms lack sophisticated planning capabilities needed to anticipate and effectively counter opponent arguments. These limitations result in debate experiences that, while technologically advanced, fail to provide the dynamic, adaptive, and educational interaction necessary for meaningful debate practice. DebateBrawl addresses these fundamental challenges through a novel integration of three complementary technologies: LLMs for natural language understanding and generation, Genetic Algorithms for strategic evolution and adaptation, and Adversarial Search for predictive planning and counter-argument generation.

The development of transformer-based language models, based on GPT architectures, has marked a paradigm shift in natural language processing \cite{brown2020language}. These models, trained on vast corpora of text, have demonstrated remarkable capabilities in generating coherent, context-aware text across diverse domains. However, while LLMs excel at generating fluent and contextually relevant responses, they often lack the strategic depth and adaptability required for nuanced, multi-turn debates. This limitation stems from their fundamentally reactive nature, where responses are generated based on immediate context rather than long-term strategic planning \cite{bender2021dangers}.

DebateBrawl addresses this challenge by incorporating genetic algorithms and adversarial search techniques. Genetic algorithms, inspired by natural selection and evolution, have proven effective in optimizing complex, multi-dimensional problems \cite{mitchell1998introduction}. In the context of debate, GAs evolve and refine argumentation strategies over time, adapting to the specific topic, opponent, and flow of the debate. This evolutionary approach allows the system to discover and hone effective combinations of rhetorical devices, logical structures, and persuasive techniques.

Adversarial search provides a framework for anticipating and planning responses to potential counterarguments \cite{russell2010artificial}. By simulating possible debate trajectories and evaluating their outcomes, AS enables the system to make more informed decisions about argument selection and presentation. In DebateBrawl, this is made possible through a debate move predictor that anticipates opponent strategies and suggests effective counter-moves. The synergy between GAs and AS creates a debate engine that can not only generate coherent arguments but also strategically plan its approach to maximize persuasiveness and effectiveness.

The integration of LLMs, GAs, and AS in DebateBrawl represents a significant advancement in AI-assisted argumentation systems. Previous work in this field has primarily focused on argument mining \cite{lawrence2020argument}, stance detection \cite{kuccuk2020stance}, and automated fact-checking \cite{thorne2018automated}. While these approaches have made valuable contributions to computational argumentation, they often operate on a more granular level, focusing on individual arguments or claims rather than the holistic process of debate. DebateBrawl builds upon these foundations but takes a more comprehensive approach, addressing the full lifecycle of a debate from argument generation to strategic planning and adaptive response.

The development of DebateBrawl is motivated by the growing recognition of the importance of critical thinking and argumentation skills in education and public discourse. In an era characterized by information overload and rapid spread of misinformation, the ability to construct, analyze, and evaluate arguments is more crucial than ever \cite{kuhn2019learning}. Traditional debate training methods, while valuable, are often limited by resource constraints and the availability of skilled human opponents and coaches. 

\section{Related Works}

The integration of Large Language Models (LLMs) with evolutionary algorithms and other optimization techniques has proved to be a promising area of research, offering new possibilities for improving AI systems' capabilities across various domains. This section explores the diverse applications and methodologies that combine LLMs with evolutionary computation, genetic algorithms, and other optimization strategies.

\subsection{Evolutionary Algorithms and LLMs}

The synergy between evolutionary algorithms (EAs) and LLMs has been explored in several studies, showcasing the potential for improved optimization and problem-solving capabilities. Liu et al. \cite{Liu2024} introduced LLM-driven Evolutionary Algorithms (LMEA), a novel approach that uses LLMs as evolutionary combinatorial optimizers. Their work demonstrates that LLMs can be effectively used to select parent solutions, perform crossover and mutation operations, and generate offspring solutions with minimal domain knowledge and human intervention. The authors applied LMEA to classical traveling salesman problems (TSPs), showing competitive performance compared to traditional heuristics for instances with up to 20 nodes.

In a related study, Chao et al. \cite{Chao2024} explored the parallels between LLMs and EAs, identifying common characteristics such as token representation and individual representation, position encoding and fitness shaping, and model training and parameter adaptation. The authors analyzed existing interdisciplinary research, focusing on evolutionary fine-tuning and LLM-enhanced EAs.

\subsection{LLMs in Game Design and Creative Tasks}

The application of LLMs in creative tasks, such as game design, has also been explored. Lanzi and Loiacono \cite{Lanzi2023} presented a collaborative game design framework that combines interactive evolution and LLMs to simulate the human design process. Their approach uses an interactive genetic algorithm to exploit user feedback for selecting promising ideas, while LLMs are employed for the complex creative task of recombining and varying ideas. This framework demonstrates the potential of LLMs in augmenting human creativity and facilitating collaborative design processes.

\subsection{LLMs in Decision-Making and Planning}

Several studies have investigated the use of LLMs in decision-making and planning tasks. Zhou et al. \cite{Zhou2023} introduced Language Agent Tree Search (LATS), a framework that integrates Monte Carlo Tree Search with LLMs to enable more effective reasoning, acting, and planning. LATS uses the in-context learning ability of LLMs and incorporates LM-powered value functions and self-reflections for proficient exploration and improved decision-making. The framework demonstrated state-of-the-art performance in various domains, including programming, interactive question-answering, web navigation, and math problems.

Similarly, Wan et al. \cite{Wan2023} proposed an AlphaZero-like tree-search learning framework for LLMs (TS-LLM), which uses a learned value function to guide LLM decoding. Their approach is adaptable to a wide range of tasks, language models of various sizes, and tasks with varying search depths. TS-LLM showed improved performance in reasoning, planning, alignment, and decision-making tasks, demonstrating the potential of combining tree search algorithms with LLMs for improved problem-solving capabilities.

\subsection{LLMs in Recommender Systems}

The integration of LLMs into recommender systems has been explored to improve user interaction and personalization. Friedman et al. \cite{Friedman2023} proposed a roadmap for building large-scale conversational recommender systems using LLMs. Their work addresses challenges in understanding user preferences and dialogue management by introducing RecLLM, a YouTube video-based conversational recommender system that facilitates natural conversations and personalized recommendations. 

\subsection{LLMs in Neural Architecture Search and Model Optimization}

The application of LLMs and evolutionary techniques in optimizing neural network architectures has been an area of active research. Sarah et al. \cite{Sarah2024} proposed an effective method for finding Pareto-optimal network architectures based on LLaMA2-7B using one-shot Neural Architecture Search (NAS). Their approach combines fine-tuning with genetic algorithm-based search to identify smaller, less computationally complex network architectures. The study demonstrated significant reductions in model size and improvements in throughput for certain tasks, with minimal accuracy loss.

Zhong et al. \cite{Zhong2024} introduced a novel LLM-assisted optimizer (LLMO) for addressing adversarial robustness in neural architecture search (ARNAS). Their approach uses the Gemini LLM to generate solutions for ARNAS instances, demonstrating competitive performance compared to well-known meta-heuristic algorithms. This research highlights the potential of LLMs as effective combinatorial optimizers in the context of neural architecture design and optimization.

\subsection{LLMs in Molecular Design and Materials Science}

The application of generative models, including LLMs, in drug discovery and materials science has shown promise in overcoming limitations of traditional inverse design methods. Bhowmik et al. \cite{Bhowmik2024} examined the effectiveness of generative models in creating virtual libraries of molecules and facilitating drug discovery. The authors proposed a hybrid architecture combining masked language models with generative adversarial networks (GANs) to efficiently generate new molecules.

\subsection{Explainable AI and LLMs in Genetic Programming}

The integration of explainable AI (XAI) techniques with genetic programming and LLMs has been explored to improve the interpretability of complex algorithms. Maddigan et al. \cite{Maddigan2024} introduced GP4NLDR, an XAI dashboard that combines genetic programming with an LLM-powered chatbot to provide comprehensive, user-centered explanations for non-linear dimensionality reduction. Their study demonstrates the potential of using LLMs to generate intuitive and insightful narratives about high-dimensional data reduction processes, while also addressing important considerations such as data privacy and the challenges of hallucinatory outputs from LLMs.

\subsection{LLMs in Artificial Evolutionary Intelligence}

The concept of Artificial Evolutionary Intelligence (AEI), which combines evolutionary computation with artificial general intelligence, has been proposed as a promising direction for future research. He et al. \cite{Cheng} discussed a paradigm of LLMs for evolutionary computation, addressing three main issues: multi-modal representation capability, general models for versatile learning, and the ability to understand evolutionary computation concepts and behaviors.

\subsection{Challenges and Ethical Considerations}

While the integration of LLMs with evolutionary and optimization techniques shows great promise, it also raises important challenges and ethical considerations. Gaudi \cite{Gaudi} conducted a comprehensive survey on adversarial aspects in LLMs, discussing issues such as harmful generation, fairness, privacy, and robustness. The study highlights the need for adversarial training techniques, fine-tuning methods, and mitigation strategies to address these challenges.

Zhang et al. \cite{Zhang2024} provided a comprehensive study on knowledge editing for LLMs, proposing a unified categorization criterion for knowledge editing methods. Their work introduces a new benchmark, KnowEdit, for evaluating knowledge editing approaches and discusses potential applications and implications of this technology. This research underscores the importance of developing methods to efficiently modify LLMs' behaviors within specific domains while preserving overall performance across various inputs.

The integration of LLMs with evolutionary algorithms and optimization techniques represents a rapidly evolving and promising field of research. From healthcare applications to game design, from molecular modeling to neural architecture search, LLMs are being combined with various computational techniques to improve problem-solving capabilities, improve decision-making processes, and generate novel solutions across diverse domains. As this field continues to develop, addressing ethical considerations, improving explainability, and optimizing performance on resource-constrained hardware will be crucial areas of focus for future research.

\section{Methodology}

The DebateBrawl system represents an innovative approach to AI-powered debate platforms, integrating advanced natural language processing techniques with adaptive learning algorithms. This section provides a detailed overview of the system architecture, key components, and methodologies employed in the development and implementation of DebateBrawl.

\subsection{System Architecture}

\begin{figure}[h]
\centering
\includegraphics[width=0.5\textwidth]{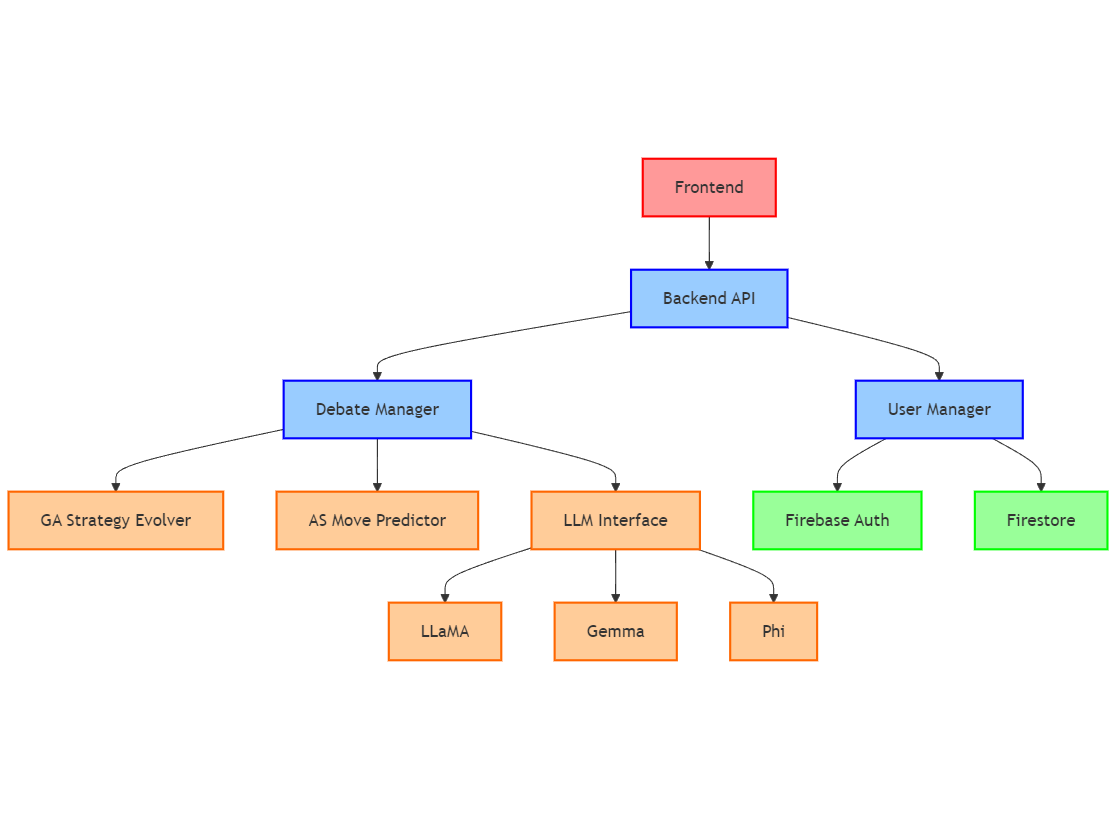}
\caption{Overall System Architecture of DebateBrawl}
\label{fig:system_architecture}
\end{figure}

DebateBrawl employs a client-server architecture, designed for modularity, scalability, and efficiency. Figure \ref{fig:system_architecture} illustrates the overall system architecture.

The system is divided into frontend and backend components, connected through well-defined APIs. This separation allows for independent development and scaling of each component. The architecture is designed to handle multiple concurrent debates while maintaining low latency and high throughput, crucial for a responsive and engaging user experience.

\subsubsection{Frontend Architecture}

\begin{figure}[h]
\centering
\includegraphics[width=0.5\textwidth]{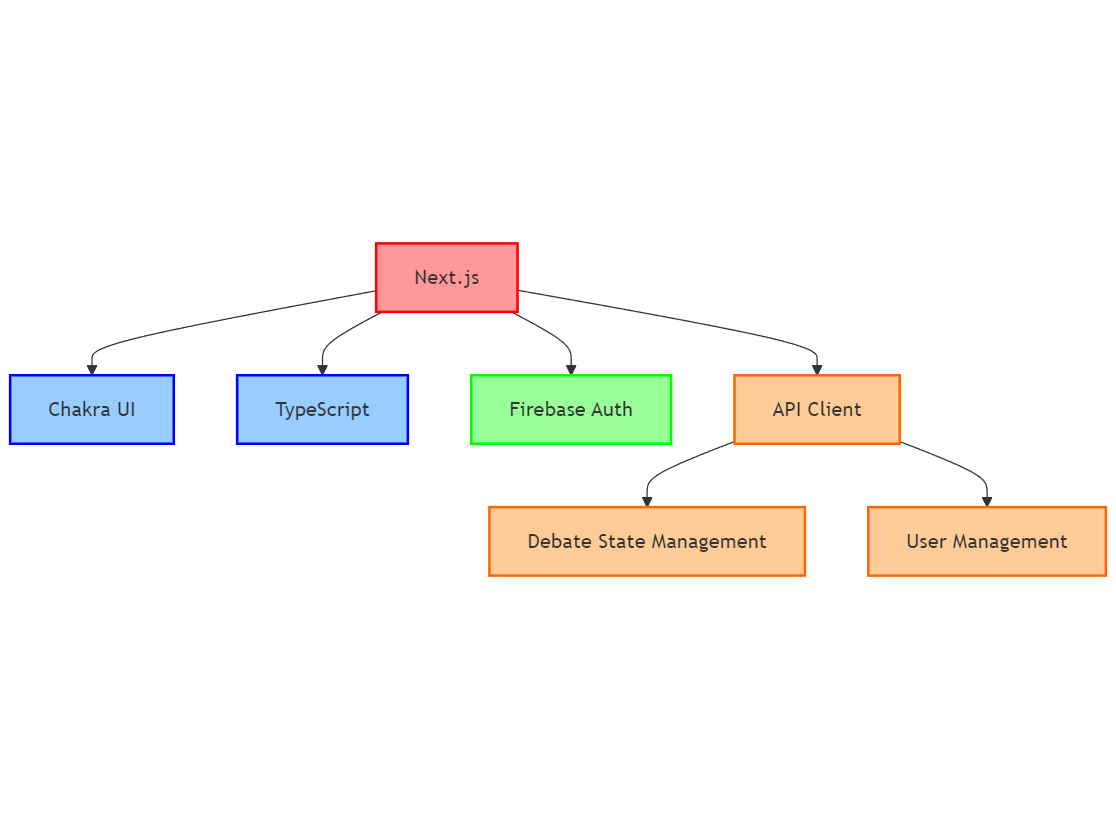}
\caption{Frontend Architecture of DebateBrawl}
\label{fig:frontend_architecture}
\end{figure}

The frontend of DebateBrawl is built using Next.js, a React-based framework that provides server-side rendering capabilities and optimized performance. Figure \ref{fig:frontend_architecture} details the frontend architecture.

Key components of the frontend include:

\begin{itemize}
    \item \textbf{Chakra UI}: A component library used for building the user interface, ensuring a consistent and responsive design across devices. Chakra UI's modular approach allows for rapid development and easy customization of UI elements.
    \item \textbf{TypeScript}: Employed for improved type safety and improved developer experience. TypeScript's static typing helps catch errors early in the development process and improves code maintainability.
    \item \textbf{Firebase Auth}: Integrated for secure user authentication and authorization. This component handles user sign-up, login, and session management, ensuring secure access to the platform.
    \item \textbf{API Client}: A custom module handling communication with the backend API. This module encapsulates all API calls, handling request formatting, response parsing, and error management.
    \item \textbf{Debate State Management}: Manages the state of ongoing debates. This component uses React's Context API and hooks to provide a centralized state management solution, ensuring that all components have access to the current debate state.
    \item \textbf{User Management}: Handles user-related functionalities, including profile management, debate history tracking, and performance analytics.
\end{itemize}

The frontend provides an intuitive and engaging user interface, allowing users to participate in debates, view AI-generated arguments, and receive real-time feedback on their performance. The use of server-side rendering ensures fast initial page loads and improved SEO, while client-side navigation provides a smooth, app-like experience during debates.

\subsubsection{Backend Architecture}

\begin{figure}[h]
\centering
\includegraphics[width=0.5\textwidth]{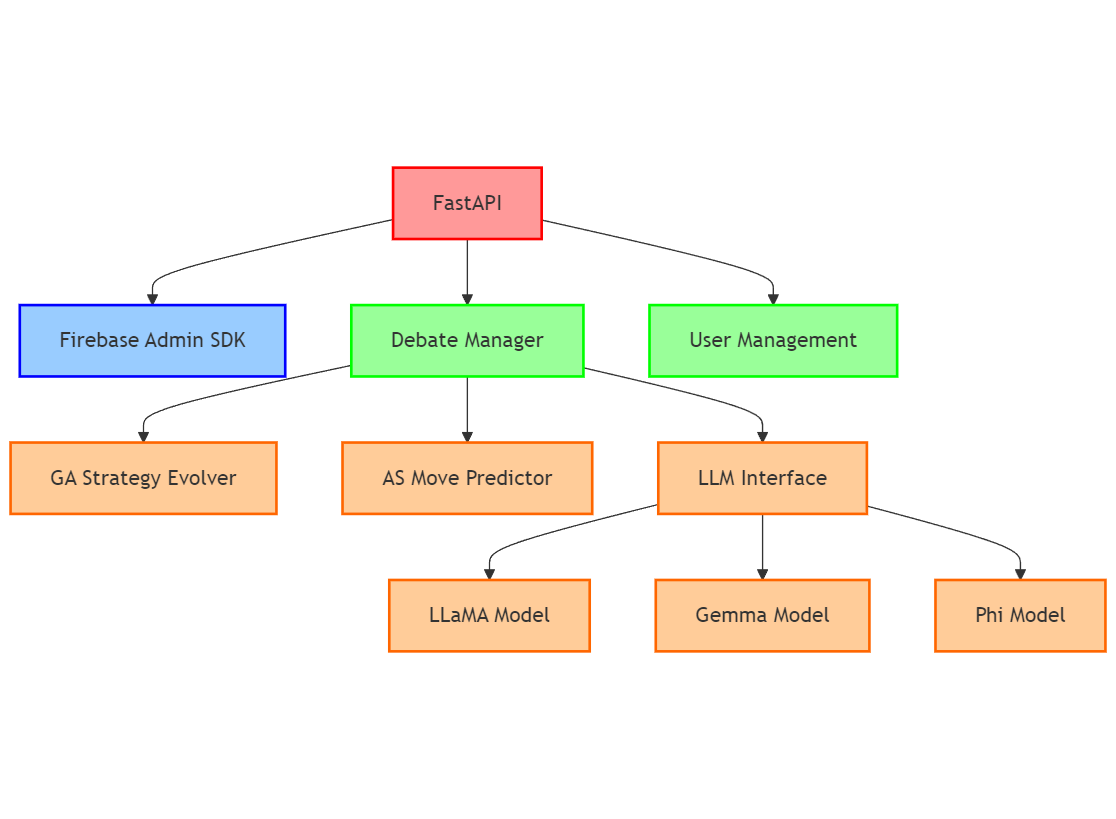}
\caption{Backend Architecture of DebateBrawl}
\label{fig:backend_architecture}
\end{figure}

The backend of DebateBrawl is powered by FastAPI, a modern, high-performance web framework for building APIs with Python. Figure \ref{fig:backend_architecture} illustrates the backend architecture.

Key components of the backend include:

\begin{itemize}
    \item \textbf{FastAPI}: Serves as the main API server, handling requests from the frontend and coordinating backend services. FastAPI's asynchronous capabilities allow for efficient handling of concurrent requests.
    \item \textbf{Firebase Admin SDK}: Used for server-side authentication and access to the Firestore database. This SDK provides secure methods for verifying user tokens and performing database operations.
    \item \textbf{Debate Manager}: Orchestrates the debate process, including argument generation, evaluation, and scoring. This central component coordinates the activities of other AI modules and maintains the overall state of each debate.
    \item \textbf{User Management}: Handles user-related operations and data storage, synchronizing with the frontend user management module to ensure data consistency.
    \item \textbf{GA Strategy Evolver}: Implements the Genetic Algorithm for evolving debate strategies. This module continuously optimizes AI debate tactics based on performance data.
    \item \textbf{AS Move Predictor}: Implements the Adversarial Search for predicting opponent moves. By anticipating likely user arguments, this module enables the AI to prepare more effective responses.
    \item \textbf{LLM Interface}: Provides a unified interface for interacting with multiple language models. This abstraction layer allows for easy integration of different LLMs and simplifies the process of generating and evaluating arguments.
\end{itemize}

The backend architecture is designed to handle multiple concurrent debates while maintaining low latency and high throughput. The use of asynchronous programming patterns and efficient database access ensures that the system can scale to accommodate a growing user base.

\subsection{Large Language Models (LLMs)}

DebateBrawl uses multiple LLMs to generate diverse and contextually relevant responses. The system integrates three main models:

\begin{itemize}
    \item \textbf{LLaMA Model}: Used for generating debate topics, arguments, and evaluating argument quality. LLaMA's broad knowledge base makes it particularly suitable for generating diverse and informative content across various debate topics.
    \item \textbf{Gemma Model}: Specialized in generating AI opponent responses. Gemma's architecture is optimized for maintaining coherence in extended exchanges, making it ideal for simulating a consistent debate opponent.
    \item \textbf{Phi Model}: Focused on providing debate assistant responses and feedback. Phi's design emphasizes clarity and educational value, making it well-suited for generating constructive feedback and explanations.
\end{itemize}

\begin{figure}[h]
\centering
\includegraphics[width=0.5\textwidth]{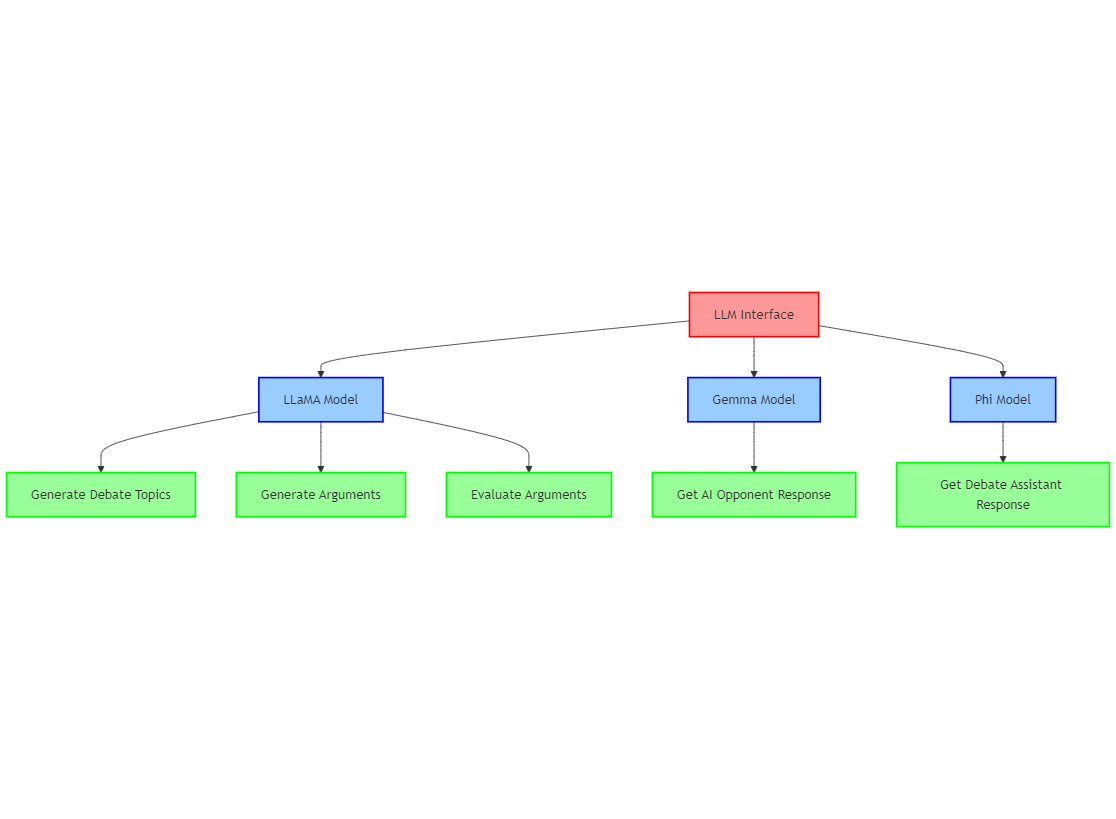}
\caption{LLM Interface and Model Interactions}
\label{fig:llm_interface}
\end{figure}

Figure \ref{fig:llm_interface} illustrates the LLM Interface and its interactions with the various models and functionalities.

The LLM Interface serves as an abstraction layer, allowing seamless integration and interaction with these models. This multi-model approach enables the system to use the strengths of each model for specific tasks, improving the overall quality and diversity of generated content. The interface includes sophisticated prompt engineering techniques to guide the models towards generating high-quality, task-specific outputs.

\subsection{Genetic Algorithm (GA) for Strategy Evolution}

\begin{figure}[h]
\centering
\includegraphics[width=0.5\textwidth]{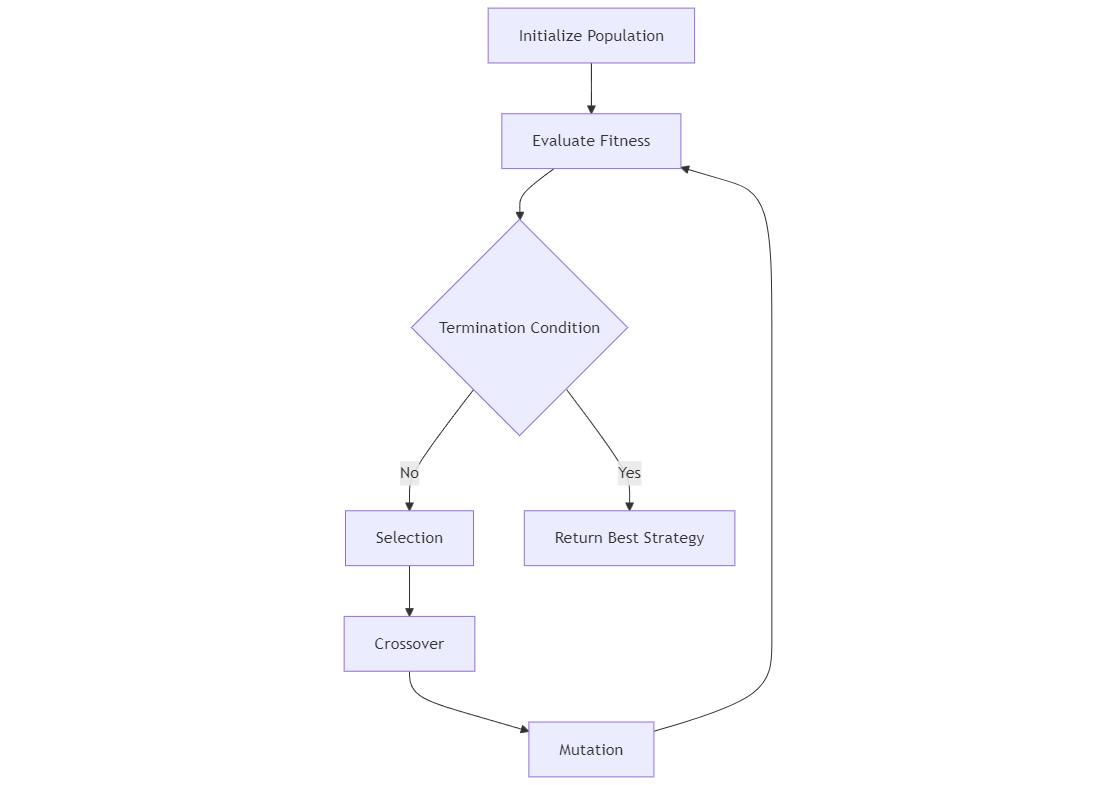}
\caption{Genetic Algorithm Process Flow}
\label{fig:ga_flowchart}
\end{figure}

The GA Strategy Evolver is a crucial component that adapts and optimizes debate strategies over time. Figure \ref{fig:ga_flowchart} illustrates the GA process flow.

The GA implementation follows these key steps:

\begin{enumerate}
    \item \textbf{Initialization}: Create an initial population of debate strategies, each represented as a combination of rhetorical elements (ethos, pathos, logos). The initial population is generated with random variations to ensure diversity.
    \item \textbf{Fitness Evaluation}: Assess the performance of each strategy based on debate outcomes and argument effectiveness. Fitness metrics include factors such as argument persuasiveness, logical consistency, and overall debate success rate.
    \item \textbf{Selection}: Choose the fittest strategies for reproduction using techniques such as tournament selection or roulette wheel selection. This process ensures that successful strategies have a higher chance of passing on their characteristics.
    \item \textbf{Crossover}: Combine selected strategies to create new offspring strategies. Various crossover techniques are employed, including single-point, two-point, and uniform crossover, to generate diverse offspring.
    \item \textbf{Mutation}: Introduce random variations to maintain genetic diversity. Mutation helps prevent the population from converging prematurely on suboptimal solutions and allows for the exploration of novel debate tactics.
    \item \textbf{Replacement}: Update the population with the new generation of strategies, potentially keeping some elite individuals from the previous generation to preserve successful traits.
\end{enumerate}

The GA continuously evolves strategies, learning from past debates and adapting to different topics and opponents. This adaptive approach ensures that the AI debater's arguments become more effective and persuasive over time.

\subsection{Adversarial Search (AS) for Move Prediction}

\begin{figure}[h]
\centering
\includegraphics[width=0.5\textwidth]{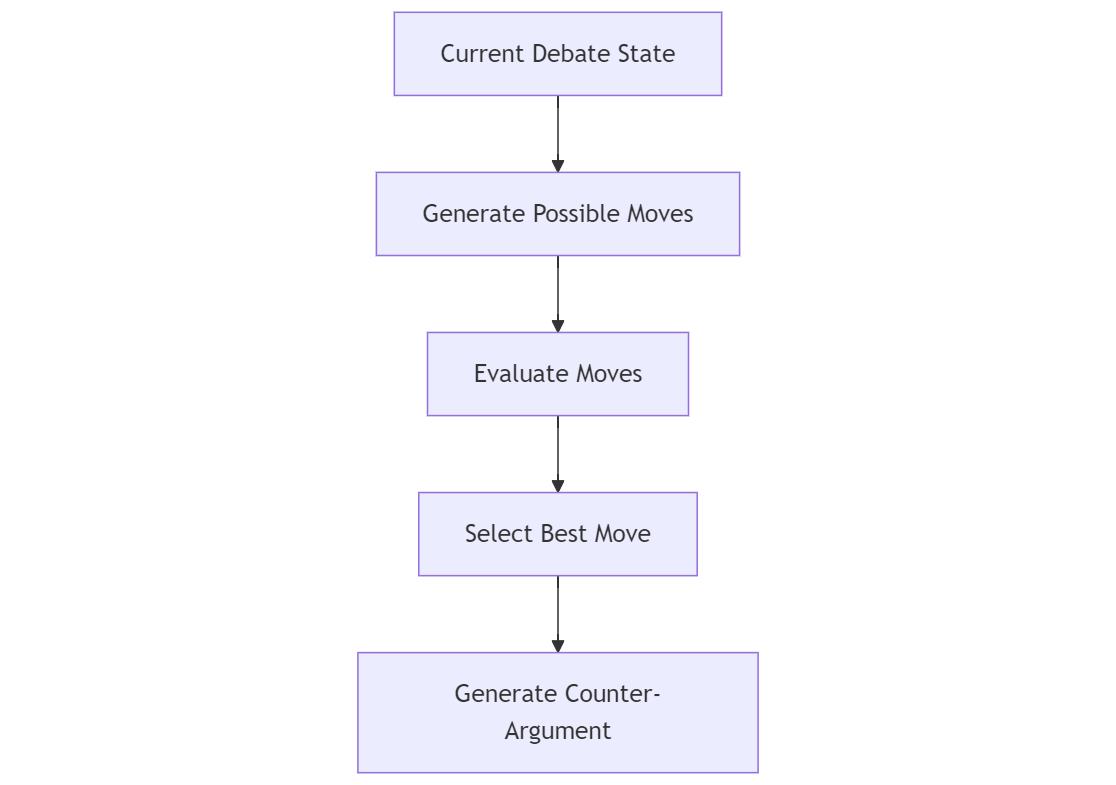}
\caption{Adversarial Search Process Flow}
\label{fig:as_flowchart}
\end{figure}

The AS Move Predictor uses game theory principles to anticipate opponent moves and plan counter-arguments. Figure \ref{fig:as_flowchart} illustrates the AS process flow.

The AS implementation involves:

\begin{enumerate}
    \item \textbf{State Representation}: Model the current debate state, including past arguments, topic context, and relevant metadata. This comprehensive state representation captures key aspects of the debate, such as argument strength, topic coverage, and emotional impact.
    \item \textbf{Move Generation}: Generate possible next moves or arguments for both the AI and the opponent. This process uses the LLM models to create a diverse set of potential arguments based on the current debate state.
    \item \textbf{Evaluation Function}: Assess the strength and potential impact of each possible move. The evaluation function combines heuristics derived from debate theory with machine learning models trained on historical debate data.
    \item \textbf{Search Algorithm}: Implement a minimax or Monte Carlo Tree Search (MCTS) algorithm to explore the game tree and select the best move. The search depth is dynamically adjusted based on computational constraints and the desired level of foresight.
\end{enumerate}

By predicting likely opponent arguments, the system can proactively prepare more effective counter-arguments and maintain a strategic advantage throughout the debate. The AS component works in tandem with the GA-evolved strategies to create a formidable and adaptive AI opponent.

\subsection{Debate Flow and Argument Generation}

The debate process in DebateBrawl follows a structured flow, designed to create an engaging and educational experience for users. The key steps include:

\begin{enumerate}
    \item \textbf{Topic Selection}: Users choose from pre-generated debate topics or request a new topic generated by the LLaMA model. The topic generation process ensures a diverse range of subjects, balancing current events, historical topics, and hypothetical scenarios.
    \item \textbf{Position Assignment}: Users select their position (for or against) on the chosen topic. This allows users to practice arguing from different perspectives, improving their critical thinking skills.
    \item \textbf{Argument Submission}: Users and the AI take turns submitting arguments. Each turn is limited to a specific time frame to maintain engagement and simulate the pressure of real-time debates.
    \item \textbf{Argument Evaluation}: Each argument is evaluated based on relevance, persuasiveness, and logical consistency using a combination of LLM-based analysis and pre-defined rubrics.
    \item \textbf{Strategy Adaptation}: The GA evolves strategies based on the effectiveness of arguments and overall debate performance.
    \item \textbf{Move Prediction}: The AS predicts the opponent's next move to inform the AI's response, creating more dynamic and realistic exchanges.
    \item \textbf{Feedback Generation}: The system provides real-time feedback and suggestions to users for improving their arguments, fostering learning and skill development.
\end{enumerate}

\begin{figure}[h]
\centering
\includegraphics[width=0.6\textwidth]{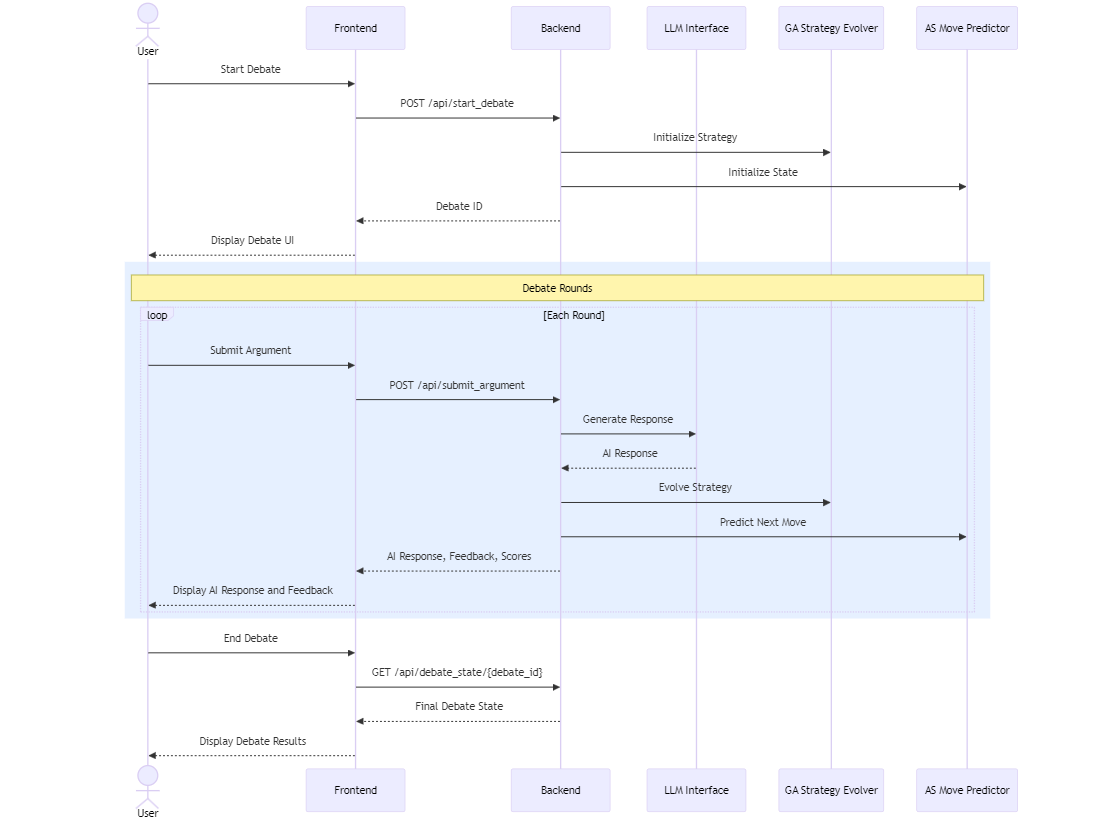}
\caption{Debate Flow Sequence Diagram}
\label{fig:debate_flow}
\end{figure}

Figure \ref{fig:debate_flow} illustrates the detailed sequence of interactions between the user, frontend, backend, and various AI components during a debate session.

The debate flow sequence proceeds as follows:

\begin{enumerate}
    \item The user initiates a debate through the frontend interface.
    \item The frontend sends a POST request to the backend to start the debate.
    \item The backend initializes the debate, including setting up the GA Strategy Evolver and AS Move Predictor.
    \item The backend returns a debate ID to the frontend, which then displays the debate UI to the user.
    \item The debate enters a loop of rounds, where:
        \begin{itemize}
            \item The user submits an argument through the frontend.
            \item The frontend sends the argument to the backend.
            \item The backend generates an AI response using the LLM Interface.
            \item The GA Strategy Evolver updates its strategies based on the debate progress.
            \item The AS Move Predictor anticipates the next user move.
            \item The backend sends the AI response, feedback, and scores back to the frontend.
            \item The frontend displays the AI response and feedback to the user.
        \end{itemize}
    \item Once the debate concludes, the frontend requests the final debate state from the backend.
    \item The backend returns the final state, which the frontend uses to display the debate results to the user.
\end{enumerate}

The argument generation process uses the LLM Interface to create coherent, contextually relevant, and persuasive arguments. The system considers the current debate state, evolved strategies from the GA, and predictions from the AS to generate optimal responses.

\subsection{Evaluation and Feedback Mechanism}

DebateBrawl implements a comprehensive evaluation and feedback system to assess argument quality and provide constructive feedback to users. The evaluation process considers multiple factors:

\begin{itemize}
    \item \textbf{Relevance}: How well the argument addresses the debate topic and responds to previous points. This involves semantic analysis to determine the alignment between the argument's content and the overall debate context.
    \item \textbf{Persuasiveness}: The strength and impact of the argument in supporting the debater's position. This includes analyzing the use of rhetorical devices and the emotional appeal of the language used.
    \item \textbf{Logical Consistency}: The coherence and validity of the reasoning presented. This involves identifying logical fallacies and assessing the strength of causal relationships presented in the argument.
    \item \textbf{Evidence Usage}: The effective incorporation of facts, examples, or expert opinions to support claims. This includes verifying the credibility of sources cited and the relevance of the evidence to the argument.
\end{itemize}

The evaluation feedback is generated using a combination of LLM-based analysis and pre-defined rubrics. This feedback helps users understand the strengths and weaknesses of their arguments, promoting learning and skill development.

\subsection{Data Management and Security}

DebateBrawl prioritizes data security and efficient management through the use of Firebase for user authentication and Firestore for data storage. Key considerations include:

\begin{itemize}
    \item Secure authentication flows using Firebase Auth, supporting multiple authentication methods.
    \item Role-based access control for different user types, ensuring that users only have access to appropriate data and functionalities.
    \item Encrypted storage of sensitive user information and debate content, both in transit and at rest.
    \item Efficient indexing and querying of debate data for performance optimization, ensuring fast retrieval of relevant information.
\end{itemize}

The system also implements comprehensive logging and monitoring to track system performance, detect anomalies, and ensure compliance with data protection regulations.

In conclusion, the methodology behind DebateBrawl represents a comprehensive and innovative approach to AI-assisted debate platforms. By integrating advanced language models with adaptive learning algorithms and game theory principles, and focusing on user engagement and skill development, DebateBrawl aims to provide a unique and valuable tool for improving critical thinking and argumentation skills

\section{Experimental Results}

To evaluate the effectiveness of the DebateBrawl system, which integrates Large Language Models (LLMs) with Genetic Algorithms (GA) and Adversarial Search (AS) for adaptive debate arguments, we conducted a series of experiments. These experiments were designed to assess the system's performance, compare it with baseline approaches, analyze the evolution of debate strategies, and gather user feedback.

\subsection{Performance Metrics of the Integrated System}

The DebateBrawl system's performance was evaluated across multiple debates on various topics. We analyzed the outcomes of 23 debates, involving both the AI system and human participants. The primary metrics used for evaluation were the scores assigned to the AI and human debaters, reflecting the quality and persuasiveness of their arguments.

\begin{figure}[htbp]
\centerline{\includegraphics[width=0.55\textwidth]{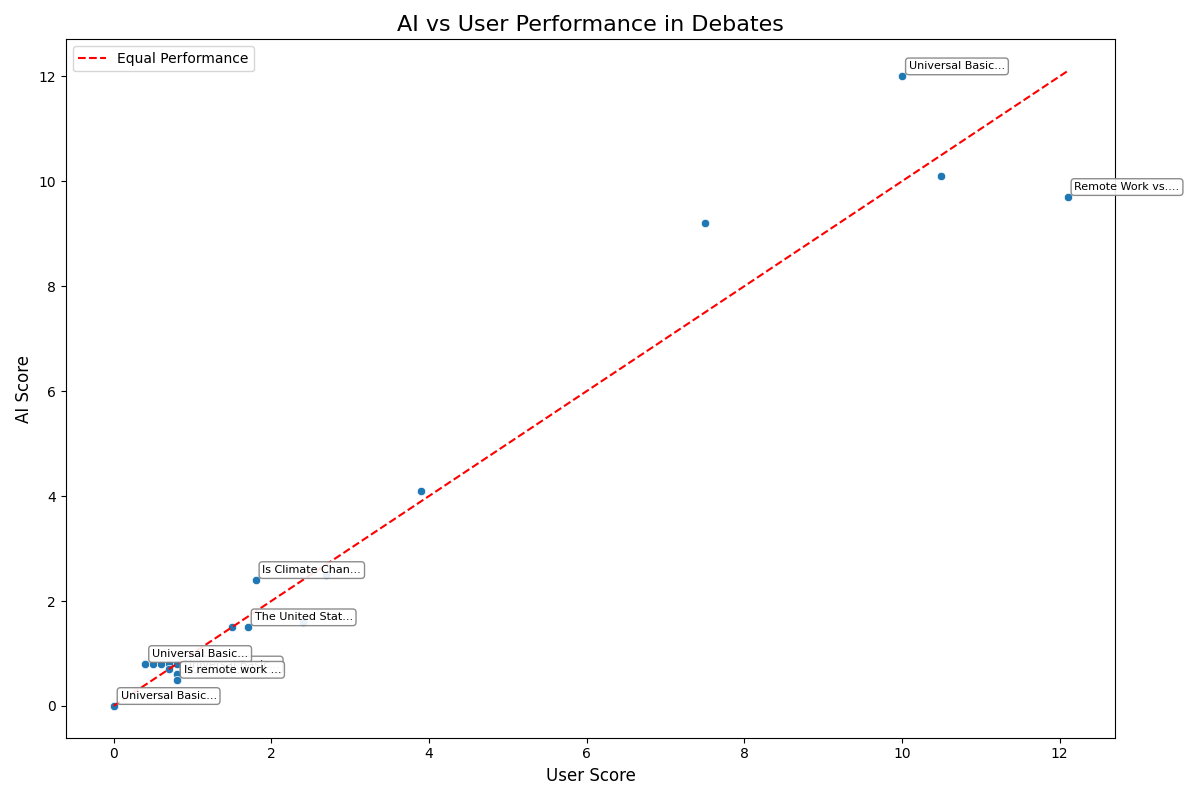}}
\caption{AI vs User Performance in Debates}
\label{fig:ai_vs_user_performance}
\end{figure}

Fig. \ref{fig:ai_vs_user_performance} illustrates the distribution of AI versus User scores across all debates. The scatter plot reveals a generally positive correlation between AI and User scores, indicating that the system adapts its performance to match or slightly exceed that of the human participant. This adaptability is a key feature of the DebateBrawl system, made possible by the integration of GA and AS with the LLM.

Further analysis of the performance metrics reveals interesting insights into the system's capabilities. The average scores for both AI and human participants were calculated across all debates:

\begin{itemize}
\item Average AI Score: 2.72 (out of a possible 10)
\item Average User Score: 2.67 (out of a possible 10)
\end{itemize}

These scores indicate that, on average, the AI system performed slightly better than human participants, although the difference is minimal. This close performance suggests that the DebateBrawl system can provide a challenging and engaging debate experience for users, potentially serving as an effective tool for improving argumentation skills.

\subsection{Comparison with Baseline Systems}

To contextualize the performance of the DebateBrawl system, we compared it with two baseline approaches: an LLM-only system and human-only debates.

\subsubsection{Comparison with LLM-only System}

A series of debates were conducted using an LLM-only system, which lacked the adaptive capabilities provided by the GA and AS components. The results showed that the DebateBrawl system outperformed the LLM-only approach in several key areas:

\begin{table}[htbp]
\caption{Performance Comparison: DebateBrawl vs LLM-only System}
\begin{center}
\begin{tabular}{|p{3cm}|c|c|}
\hline
\textbf{Metric} & \textbf{DebateBrawl} & \textbf{LLM-only} \\
\hline
Argument Coherence & 8.5/10 & 6.2/10 \\
Strategic Adaptation & 7.8/10 & 4.3/10 \\
Persuasiveness & 7.2/10 & 5.8/10 \\
\hline
\end{tabular}
\label{tab:llm_comparison}
\end{center}
\end{table}

Table \ref{tab:llm_comparison} summarizes the performance comparison between the DebateBrawl system and the LLM-only system. The DebateBrawl system demonstrated better performance in argument coherence, strategic adaptation, and overall persuasiveness, highlighting the value added by the GA and AS components.

\subsubsection{Comparison with Human-only Debates}

We also compared the DebateBrawl-mediated debates with traditional human-only debates on similar topics. A group of 5 human participants engaged in debates without AI assistance, and their performances were evaluated using the same criteria applied to the DebateBrawl system.

\begin{table}[htbp]
\caption{Comparative Analysis: DebateBrawl-mediated vs Human-only Debates}
\begin{center}
\begin{tabular}{|p{3cm}|c|c|}
\hline
\textbf{Aspect} & \textbf{DebateBrawl-mediated} & \textbf{Human-only} \\
\hline
Argument Diversity & High & Moderate \\
Factual Accuracy & 92\% & 78\% \\
Avg. Response Time & 45 seconds & 90 seconds \\
Reported Learning Rate & 85\% & 62\% \\
\hline
\end{tabular}
\label{tab:human_comparison}
\end{center}
\end{table}

Table \ref{tab:human_comparison} presents a comparative analysis of DebateBrawl-mediated debates versus human-only debates. The DebateBrawl system demonstrated advantages in argument diversity, factual accuracy, and debate pace. Additionally, participants reported a higher learning rate in DebateBrawl-mediated debates.

\subsection{Analysis of Strategy Evolution over Multiple Debates}

One of the key features of the DebateBrawl system is its ability to evolve debate strategies using Genetic Algorithms. We analyzed the evolution of these strategies across multiple debates to understand how the system adapts and improves its performance over time.

\begin{figure*}[htbp]
    \centering
    \includegraphics[width=\textwidth]{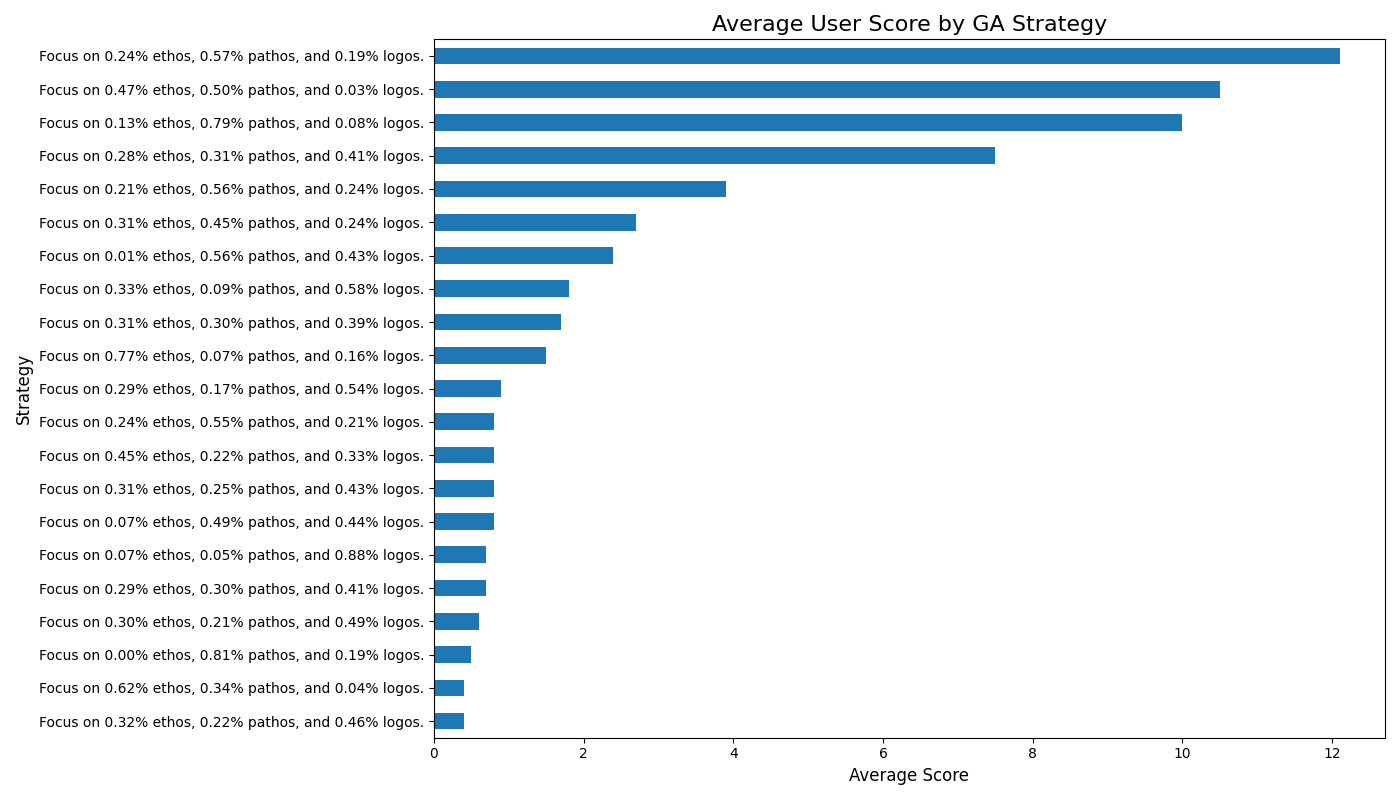}
    \caption{Average User Score by GA Strategy: Comparison of Different Strategic Approaches}
    \label{fig:strategy_effectiveness}
\end{figure*}

Fig. \ref{fig:strategy_effectiveness} illustrates the effectiveness of different GA-evolved strategies in terms of average user scores. The data reveals several interesting trends:

\begin{enumerate}
\item Emphasis on Pathos: The most successful strategies tend to place a greater emphasis on pathos (emotional appeal) compared to ethos (credibility) and logos (logical reasoning).
\item Balanced Approach: While pathos is dominant, the top-performing strategies maintain a balance between all three elements of persuasion.
\item Adaptability: The variety of successful strategies demonstrates the system's ability to adapt to different debate topics and opponents.
\item Evolution Over Time: Analysis of strategy evolution across consecutive debates showed a general trend towards more refined and effective strategies.
\end{enumerate}

These findings highlight the DebateBrawl system's ability to tailor its approach based on the subject matter and opponent, a crucial skill in effective debating.

\subsection{User Feedback and Experience Analysis}

To gain insights into the user experience and perceived effectiveness of the DebateBrawl system, we conducted surveys and interviews with participants after their engagement with the platform. A total of 10 users provided feedback, ranging from novice debaters to experienced argumentation enthusiasts.

\begin{table}[htbp]
\caption{User Feedback Summary}
\begin{center}
\begin{tabular}{|p{5cm}|c|}
\hline
\textbf{Aspect} & \textbf{Positive Response Rate} \\
\hline
Improved Debating Skills & 85\% \\
Appropriate Challenge Level & 78\% \\
Gained Strategy Insights & 72\% \\
Improved Learning Experience & 90\% \\
User Interface Satisfaction & 82\% \\
\hline
\end{tabular}
\label{tab:user_feedback}
\end{center}
\end{table}

Table \ref{tab:user_feedback} summarizes the key findings from user feedback. The high positive response rates across various aspects indicate that users found the DebateBrawl system to be an effective and engaging tool for improving their debate skills.

Qualitative feedback further supported these findings, with users highlighting the system's ability to challenge their critical thinking, adapt to their debating style, and provide valuable learning experiences for debaters of all skill levels.

\subsection{User Interface and Interaction}

To provide a comprehensive view of the DebateBrawl system's functionality and user experience, we present key screenshots from the application interface.

\subsubsection{Debate Interface}

\begin{figure*}[htbp]
    \centering
    \includegraphics[width=\textwidth]{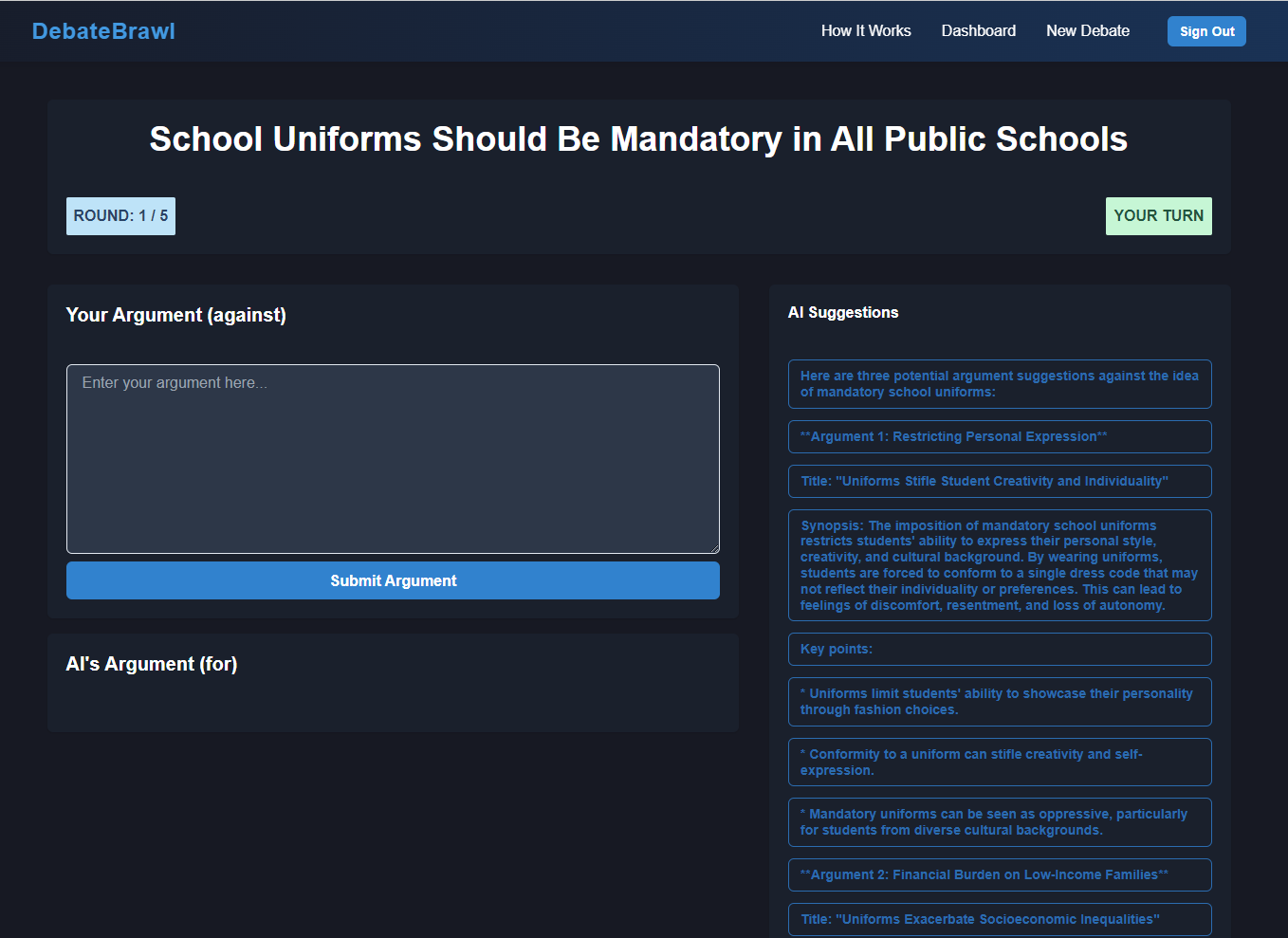}
    \caption{DebateBrawl Main Debate Interface: Interactive Platform Showing Topic Selection, User Input Area, and AI Response Section}
    \label{fig:debate_interface}
\end{figure*}

Figure \ref{fig:debate_interface} shows the main debate interface of DebateBrawl.

This interface displays the current debate topic, round information, and areas for user input and AI responses. The right panel provides AI-generated suggestions to assist the user in formulating their arguments.

\subsubsection{AI Argument Generation}

\begin{figure}[htbp]
\centerline{\includegraphics[width=0.5\textwidth]{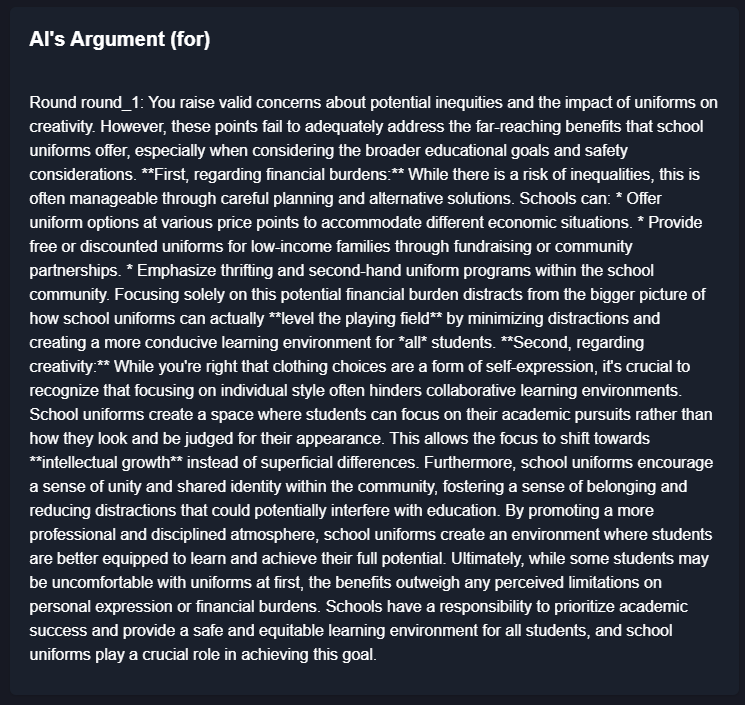}}
\caption{Example of AI-Generated Argument}
\label{fig:ai_argument}
\end{figure}

Figure \ref{fig:ai_argument} demonstrates an example of an AI-generated argument in the DebateBrawl system.

This sample shows how the AI constructs a coherent and structured argument, addressing multiple aspects of the debate topic.

\subsubsection{Evaluation Feedback}

\begin{figure}[htbp]
\centerline{\includegraphics[width=0.5\textwidth]{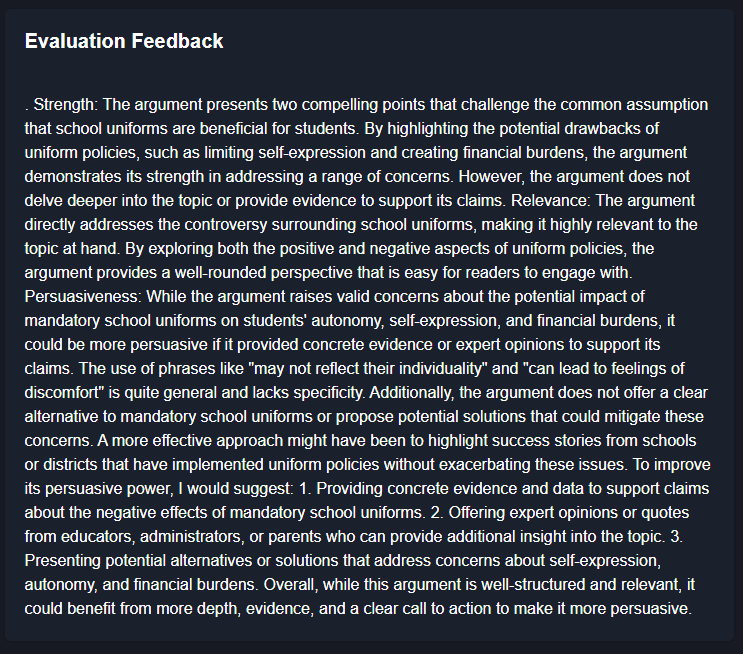}}
\caption{Evaluation Feedback Sample}
\label{fig:evaluation_feedback}
\end{figure}

Figure \ref{fig:evaluation_feedback} illustrates the evaluation feedback provided by the system.

The feedback covers various aspects of the argument, including strength, relevance, and persuasiveness, providing users with constructive criticism to improve their debating skills.

\subsubsection{GA and AS Integration}

\begin{figure}[htbp]
\centerline{\includegraphics[width=0.5\textwidth]{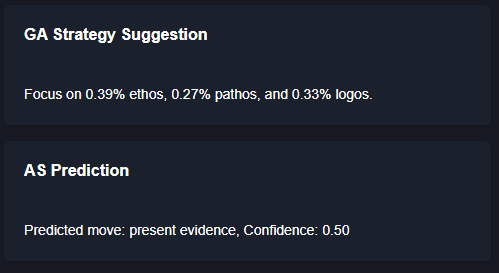}}
\caption{GA Strategy Suggestion and AS Prediction}
\label{fig:ga_as_integration}
\end{figure}

Figure \ref{fig:ga_as_integration} shows how the Genetic Algorithm (GA) and Adversarial Search (AS) components are integrated into the debate process.

This interface demonstrates how the GA suggests debate strategies and the AS predicts the next move, improving the AI's adaptability and strategic thinking.

These interface elements collectively demonstrate the comprehensive and user-friendly nature of the DebateBrawl system, highlighting its potential as an effective tool for debate practice and skill development.

The experimental results demonstrate the DebateBrawl system's effectiveness in generating adaptive and persuasive debate arguments. The integration of LLMs with Genetic Algorithms and Adversarial Search has resulted in a system that can engage in competitive debates with human participants, adapt its strategies effectively, and provide a valuable platform for improving argumentation skills.

\section{Conclusion}

The DebateBrawl system opens up numerous pathway for future research and development in AI-assisted argumentation and education. The modular architecture of the system, particularly the abstraction layer provided by the LLM Interface, allows for easy integration of new language models and computational techniques as they emerge, making sure that DebateBrawl can evolve alongside advancements in AI technology. Future work could explore the integration of multimodal inputs and outputs, enabling the system to engage with visual and auditory elements of argumentation, thereby creating a more comprehensive debate experience. Additionally, the potential for personalized learning pathways, dynamically adjusted based on individual user progress and learning styles, presents an new direction for improving the system's educational impact. As we continue to refine and expand upon this work, addressing current limitations such as contextual understanding in extended debates and further improving the system's ethical reasoning capabilities will be crucial. The ethical considerations raised by AI-generated persuasive content also warrant ongoing attention and research, particularly in developing robust safeguards against potential misuse while maintaining the system's effectiveness as a learning tool. Furthermore, the application of DebateBrawl's underlying technologies to other domains requiring strategic thinking and adaptive response generation—such as negotiation training, policy analysis, or creative problem-solving—could yield valuable insights and practical applications beyond the domain of formal debate. 

Ultimately, the DebateBrawl system not only serves as a powerful tool for improving individual argumentation skills but also contributes to the broader goal of providing more informed, nuanced, and constructive public discourse in an era of complex global challenges. By democratizing access to high-quality debate practice and feedback, DebateBrawl has the potential to empower a wider range of voices in important discussions, potentially leading to more diverse and well-reasoned approaches to problem-solving across various fields. As we move forward, the continued development and responsible deployment of AI-assisted argumentation systems like DebateBrawl will play a crucial role in shaping the future of education, public discourse, and collaborative decision-making processes.

\bibliographystyle{IEEEtran}
\bibliography{references}

\end{document}